\pdfoutput=1

\documentclass[11pt]{article}

\usepackage{coling}

\usepackage{times}
\usepackage{latexsym}
\usepackage{pifont}
\usepackage[T1]{fontenc}

\usepackage[utf8]{inputenc}
\usepackage{booktabs}
\usepackage{microtype}
\usepackage{xcolor}      
\usepackage{inconsolata}
\usepackage{multirow}
\usepackage{graphicx}
\usepackage{arydshln}

\definecolor{myyellow}{RGB}{235,184,57}
\definecolor{myred}{RGB}{240,73,110}
\definecolor{mygreen}{RGB}{12,215,159}
\title{Counting-Stars (\Pisymbol{pzd}{72}):\\ A Multi-evidence, Position-aware, and Scalable Benchmark for \\Evaluating Long-Context Large Language Models}

\author{Mingyang Song, Mao Zheng,  Xuan Luo\\
	Tencent Hunyuan \\
	{\tt nickmysong@tencent.com} \\
}

\begin{document}
	\maketitle
	\begin{abstract}
		Despite recent efforts to develop large language models with robust long-context capabilities, the lack of long-context benchmarks means that relatively little is known about their performance. To alleviate this gap, in this paper, we propose \textbf{Counting-Stars}, a multi-evidence, position-aware, and scalable benchmark designed to evaluate the multi-evidence retrieval capabilities of long-context LLMs. \textbf{Counting-Stars} comprises two counting-based multiple pieces of evidence retrieval sub-tasks: searching and reasoning. Using Counting-Stars, we conduct experiments to evaluate several long-context LLMs, including GPT-4 Turbo, Gemini 1.5 Pro, Claude3 Opus, GLM-4, and Moonshot-v1. Extensive experimental results demonstrate that Gemini 1.5 Pro achieves the best overall results, while GPT-4 Turbo exhibits the most stable performance across various tasks. Furthermore, our analysis of these LLMs, which have been extended to handle long-context scenarios, indicates that significant room for improvement remains as the length of the input context and the complexity of the tasks increase. The code and data are publicly accessible here\footnote{\url{https://github.com/nick7nlp/Counting-Stars}}.
	\end{abstract}
	
	\section{Introduction}
	
	Large language models (LLMs) have demonstrated exceptional performance across a wide range of Natural Language Processing (NLP) downstream tasks \cite{survey}. A context window of 128K tokens is crucial for LLMs and enables LLMs to perform tasks that are significantly beyond the existing paradigm, such as multi-document question answering \cite{Caciularu23}, repository-level code understanding \cite{Bairi23}, etc. An increasing number of studies focus on extending the context window these models can handle to enable LLMs to support more intricate and diverse applications. Despite these developments, the efficacy of models in long-context settings still needs to be examined, primarily due to the lack of a robust evaluation benchmark \cite{L-Eval, lost, Scaling128k}.

	In contrast to the rapid evolution of the supported context length of LLMs, existing benchmarks have lagged behind \cite{LV-Eval}. Meanwhile, it is worth mentioning that tasks in existing benchmarks are primarily short-context tasks, which only require LLMs to find evidence for answering questions within a short context to test the performance of LLMs instead of a long context \cite{LooGLE, Scaling128k}. A few benchmarks have been proposed for evaluating long-context LLMs, including LongBench \cite{longbench}, LooGLE \cite{LooGLE}, $\infty$Bench \cite{inftybench}, which have been instrumental in evaluating the performance of long-context LLMs. 
	
	Recently, the needle-in-a-haystack benchmark\footnote{\url{https://github.com/gkamradt/LLMTest_NeedleInAHaystack}} has become a popular benchmark for evaluating whether LLMs have the capability of acquiring information from long documents. Specifically, this benchmark requires LLMs to precisely retrieve a specific sentence inserted at an arbitrary position within a long context, thereby assessing their capability to search for the sentence in long contexts. However, many newly released long-context LLMs have adopted the needle-in-a-haystack benchmark to evaluate their long-context processing capabilities and have achieved nearly perfect performance. This renders the needle-in-a-haystack benchmark insufficient to distinguish the differences between these models. Not only does this demonstrate the advancements in recent long-context LLMs, but it also indicates that the needle-in-a-haystack benchmark is too simplistic to further test their capabilities. Generally, acquiring information should be the most fundamental capability of long-text LLMs, which is the prerequisite for completing complex tasks. However, existing long-context benchmarks rarely focus on the multi-evidence collection ability of LLMs. Even when they do, the amount of evidence to be collected is relatively small and not proportional to the amount of evidence that a long document should contain.

	To mitigate the shortcomings of existing benchmarks, in this paper, we propose a multi-evidence, position-aware, and scalable benchmark for evaluating long-context LLMs, named Counting-Stars. As the name suggests, the Counting-Stars refers to asking LLMs to count the numbers of stars from multiple sentences describing the number of stars counted by the little penguin inserted in the long context and then summarize into a specified answer. Through the Counting-Stars, we expect to evaluate the long context capabilities of multi-evidence searching and multi-evidence reasoning of LLMs. 
	More specifically, the former focuses on testing the capability of LLMs to retrieve evidence at different positions within the long context, which can more clearly reflect the quality of long-context modeling. When collecting evidence, reasoning is often required to ensure that the evidence gathered supports the correct answer to the question. Therefore, the latter evaluates the LLM's ability to filter out noise or incorrect information when retrieving information and the model's reasoning ability at different positions within the long context. Generally speaking, the latter is definitely more challenging than the former. In other words, the former can be treated as making LLMs distinguish between long contexts and inserted sentences (similar to the needle-in-a-haystack benchmark), while the latter involves distinguishing evidence within each inserted sentence.
	
	Experiments show that the tested LLMs can perform well on the Counting-Stars when the context length is below 32K in most cases. However, as the context length increases, the performance of all models declines. However, this decline is not absolute, meaning a model might achieve better results at 120K than at 100K. Generally, Gemini 1.5 Pro achieved the best results on all tasks, and the performance of GPT-4 Turbo is the most stable across all tasks in the Counting-Stars. Although our experiments may not fully support the loss-in-the-middle phenomenon, it can be observed that most LLMs are good at collecting the numbers of stars located at the beginning and end slightly better than those located in the middle of the long context.
	
	\begin{table*}[t]
		\centering
		\scriptsize
			\resizebox{0.99\linewidth}{!}{
				\begin{tabular}[t]{@{}lp{0.8\linewidth}@{}}
					\toprule
					\textbf{Task Name}  & \textbf{Test Example} \\
					\midrule
					\begin{tabular}[t]{@{}l@{}}
						
						Long-Context \\Multi-evidence Searching\\\textit{English Version}
						
					\end{tabular} & 
					\begin{tabular}[t]{@{}p{\linewidth}@{}}
						November 2005In the next few years, venture capital funds will find themselves squeezed from four directions.  They're already stuck with a seller's market, because of the huge amounts they raised at the end of the Bubble and still haven't invested.  This by itself is not the end of the world.  In fact, it's just a more extreme version of the norm in the VC business: too much money chasing too few deals.Unfortunately, those few deals now want less and less money, because it's getting so cheap to start a startup ... 
						
						\textbf{\textit{The little penguin counted \{number1\} \Pisymbol{pzd}{72}}}
						
						... Moore's law, which makes hardware geometrically closer to free; the Web, which makes promotion free if you're good; and better languages, which make development a lot cheaper.When we started our startup in 1995, the first three were our biggest expenses.  We had to pay \$5000 for the Netscape Commerce Server, the only software that then supported secure http connections ...
						
						\textbf{\textit{The little penguin counted \{number2\} \Pisymbol{pzd}{72}}}
						
						... people throw away computers more powerful than our first server ...
						
						......

						\textit{\textbf{On this moonlit and misty night, the little penguin is looking up at the sky and concentrating on counting \Pisymbol{pzd}{72}. Please help the little penguin collect the number of \Pisymbol{pzd}{72}, for example: {"little\_penguin": [x, x, x,...]}. The summation is not required, and the numbers in [x, x, x,...] represent the counted number of \Pisymbol{pzd}{72} by the little penguin. Only output the results in JSON format without any explanation.}}
						
						\\ 
					\end{tabular}
					\\
					\midrule
					\begin{tabular}[t]{@{}l@{}}
						
						Long-Context \\Multi-evidence Reasoning\\\textit{English Version}
						
					\end{tabular} & 
					\begin{tabular}[t]{@{}p{\linewidth}@{}}
						November 2005In the next few years, venture capital funds will find themselves squeezed from four directions.  They're already stuck with a seller's market, because of the huge amounts they raised at the end of the Bubble and still haven't invested.  This by itself is not the end of the world.  In fact, it's just a more extreme version of the norm in the VC business: too much money chasing too few deals.Unfortunately, those few deals now want less and less money, because it's getting so cheap to start a startup ... 
						
						\textbf{\textit{The little penguin counted \{wrong number1\} \Pisymbol{pzd}{72}, but found that a mistake had been made, so the counting was done again, and this time \{number1\} \Pisymbol{pzd}{72} was counted correctly.}}
						
						... Moore's law, which makes hardware geometrically closer to free; the Web, which makes promotion free if you're good; and better languages, which make development a lot cheaper.When we started our startup in 1995, the first three were our biggest expenses.  We had to pay \$5000 for the Netscape Commerce Server, the only software that then supported secure http connections ...
						
						\textbf{\textit{The little penguin counted \{wrong number2\} \Pisymbol{pzd}{72}, but found that a mistake had been made, so the counting was done again, and this time \{number2\} \Pisymbol{pzd}{72} was counted correctly.}}
						
						... people throw away computers more powerful than our first server ......

						......
						
						\textit{\textbf{On this moonlit and misty night, the little penguin is looking up at the sky and concentrating on counting \Pisymbol{pzd}{72}. Please help the little penguin collect the correct number of \Pisymbol{pzd}{72}, for example: {"little\_penguin": [x, x, x,...]}. The summation is not required, and the numbers in [x, x, x,...] represent the correctly counted number of \Pisymbol{pzd}{72} by the little penguin. Only output the results in JSON format without any explanation.}}

						\\ 
					\end{tabular}
					\\
					\bottomrule
			\end{tabular}}
			\caption{Prompt templates for the two counting tasks in the English version of Counting-Stars.}
			\label{description}
		\end{table*}
		
		\section{Counting-Stars (\Pisymbol{pzd}{72})}
		LLMs have shown remarkable performance across diverse NLP tasks but are constrained by their small context window size (short-context). Recently, various studies have expanded the context length to accommodate up to 128K tokens and more (long-context). The main difference between short- and long-context scenarios is that in the latter, LLMs need to process more information at once, which may lead to the loss of key information, resulting in decreased performance. Therefore, in long-context scenarios, the evaluation of LLMs should focus on the capability of LLMs to acquire information and distinguish incorrect information while acquiring that information.
		
		\noindent\textbf{Multi-evidence}. In long-context scenarios, answering a question may require gathering substantial evidence from different positions within the lengthy context. Therefore, it is necessary to verify the ability of LLMs to collect a large amount of evidence from a long context all at once. Moreover, to the best of our knowledge, \textit{Counting-Stars is the first long-context benchmark to significantly increase the number of pieces of evidence} (i.e., increase to $32$, $64$, $128$, $256$, $512$, and even $1024$).
		
		\noindent\textbf{Position-aware}. In long-context scenarios, a typical bad case is that when the answer to a question appears in different positions of the long context, the performance of LLMs varies greatly, such as the \textit{lost-in-the-middle} phenomenon \cite{lost}. Therefore, when evaluating the long-context LLMs, it is necessary to reveal which specific positions of evidence are missing or reasoning incorrectly through the evaluation results to analyze the problem more precisely and meticulously.
		
		\noindent\textbf{Scalable}. As mentioned earlier, developing long-context benchmarks often lags behind the speed of long-context LLMs. At the same time, constructing a long-context benchmark is difficult and expensive, so easy scalability is essential.

		In general, the capacity for a long-context LLM to do human textual instructions largely depends on its \textit{multi-evidence searching} ability. Moreover, an indispensable ability of LLMs extends beyond mere essential evidence collection to encompass \textit{reasoning} based on the collected evidence.
		Therefore, the Counting-Stars mainly evaluates the long-context capability of LLMs from two perspectives, i.e., \textit{long-context multi-evidence searching} and \textit{long-context multi-evidence reasoning}.
		Expressly, this test can be understood as asking LLMs to find and remember all the sentences in the long text that describe the little penguin counting stars and organize them into a list to return the final answer. All sentences are prior inserted into a long text at the same interval. In addition, the used long text can be any text data that is not related to the sentences describing the little penguin counting stars, such as The Story of the Stone\footnote{\textit{The Story of the Stone}, is an 18th-century Chinese novel authored by Cao Xueqin, considered to be one of the Four Great Classical Novels of Chinese literature.} for the Chinese version of the Counting-Stars and Paul Graham Essays\footnote{The English context data used in this paper is similar to the needle-in-a-haystack.} for the English version of the Counting-Stars. Next, we introduce the Counting-Stars  in detail.

		\begin{figure*}[!h]
			\centering
			\includegraphics[scale=0.39]{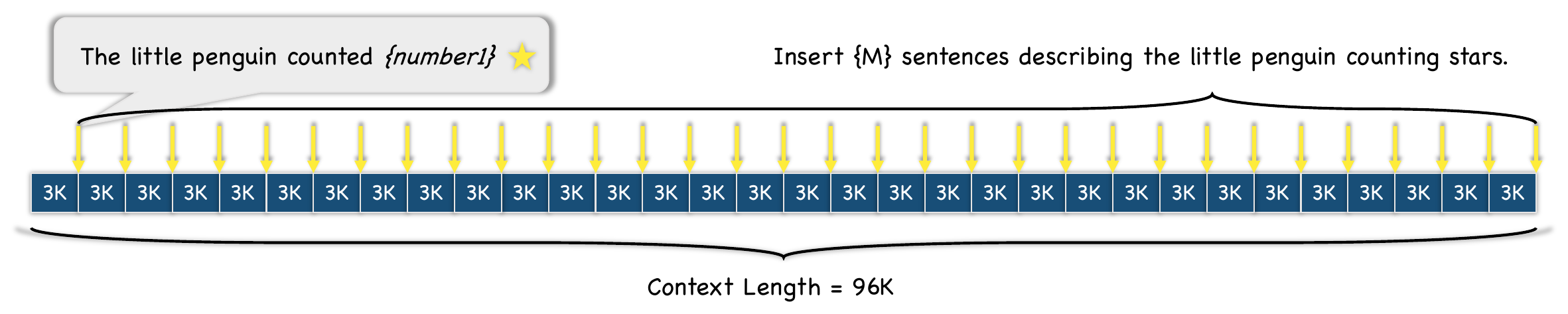}
			\caption{Illustration of how to scatter stars into the long context with the length of 96K.}
			\label{prompt}
		\end{figure*}
		
		\subsection{Long-Context Multi-evidence Searching}
		
		Multi-evidence searching refers to the capability of distinguishing and collecting critical information framed within intricate and long textual data, which bottlenecks the performance of LLMs in synthesizing contextualized knowledge to execute various tasks, from answering multi-document questions to executing complex human instructions. Furthermore, maintaining a comprehensive and accurate grasp of the input text becomes increasingly challenging as the context length increases. Therefore, in the Counting-Stars test, the first task is to examine the multi-evidence searching ability of long-context LLMs, as illustrated in Table~\ref{description} (named as \textit{Long-Context Multi-evidence Searching}). In multi-evidence searching, all sentences describing the little penguin counting stars are designed as "\textbf{\textit{The little penguin counted \{number1\} \Pisymbol{pzd}{72}}}". Here, \textit{\{number1\}} indicates the number of stars the little penguin counted. Concretely, we randomly generated all the numbers of stars as \textit{\{number1, number2, ...\}} because we found that LLMs easily slack off if a sequence of numbers is continuous or regular. In this task, we hope that LLMs collect all the numbers of stars the little penguin counted and list all digits rather than summation.
		
		\subsection{Long-Context Multi-evidence Reasoning}
		In many real-world tasks, when answering questions under a long and intricate context, it is not only necessary to collect multi-evidence information but also to reason and identify each original piece of evidence before acquiring it to avoid collecting wrong evidence. Therefore, in the Counting-Stars test, the second task is to examine the multi-evidence reasoning ability of long-context LLMs, as illustrated in Table~\ref{description} (named as \textit{Long-Context Multi-evidence Reasoning}). In multi-evidence reasoning, all sentences describing the little penguin counting stars are designed as "\textbf{\textit{The little penguin counted \{wrong number1\} \Pisymbol{pzd}{72}, but found that a mistake had been made, so the counting was done again, and this time \{number1\} \Pisymbol{pzd}{72} was counted correctly.}}". Here, \textit{\{wrong number1\}} denotes the number of stars the little penguin counted incorrectly, and \textit{\{number1\}} indicates the number of stars the little penguin counted correctly. Specifically, \textit{\{number1, 2, ...\}} are the same as the first task, and \textit{\{wrong number1, 2, ...\}} are randomly added or subtracted by one based on the \textit{\{number1, 2, ...\}}. In this task, we hope that LLMs collect all the correct numbers of stars the little penguin counted and summarize them in a list.

		\begin{table}[!]
			\scriptsize
			\centering
			\renewcommand\tabcolsep{7pt}
			\renewcommand\arraystretch{1.5}
			\begin{tabular}{lcc}
				\hline\hline
				\textsc{LLMs}        &  \textsc{Length Limit}  &  \textsc{Service Used} \\ \hline
				\textsc{GPT-4 Turbo}        &    &   \\ 
				\quad \textit{gpt4-1106-preview}        &  128K  &   Accessed from API\\ 
				\quad \textit{gpt4-0125-preview}        &  128K  &   Accessed from API\\ \cdashline{1-3}[1.5pt/2pt]
				\textsc{Gemini 1.5 Pro}        &  1M  &   Accessed from poe.com\\  \cdashline{1-3}[1.5pt/2pt]
				\textsc{Claude 3}        &    &  \\ 
				\quad \textsc{Opus}        &  200K  &   Accessed from poe.com \\ 
				\quad \textsc{Sonnet}        &  200K  &   Accessed from poe.com \\ 
				\quad \textsc{Haiku}        &  200K  &   Accessed from poe.com \\  \cdashline{1-3}[1.5pt/2pt]
				\textsc{GLM-4}        &  128K  &    Accessed from API\\  \cdashline{1-3}[1.5pt/2pt]
				\textsc{Moonshot-v1}        &  128K  &   Accessed from API\\ 
				\hline\hline   
			\end{tabular}
			\caption{LLMs used in our experiment.}
			\label{ll}
		\end{table}
		
		\subsection{Scalable Test Setting}
		
		Various approaches have been proposed to expand the context window of LLMs to accommodate even up to $128K$ input tokens or more. As the length of the context that LLMs accommodate increases, it becomes increasingly difficult to construct a qualified benchmark to evaluate them because the testing length of benchmarks can hardly be arbitrarily scaled in size. In contrast, the testing length of the Counting-Stars test can be set arbitrarily, which can be $128K$, $200K$, or even $1M$. At the same time, the amount of evidence to be collected can also be set arbitrarily. For the number of evidence, we initially set it to $M=32$, which represents the number of sentences inserted into the long context. It is worth noting that we can also set $M$ to $64$, $128$, $256$, $512$, or even $1024$. However, we find that when $M=32$, the Counting-Stars test is already difficult for many LLMs, so this paper only shows the results of each LLM when $M=32$.
		
		Another parameter that must be specially declared is the number of test samples ($N$). Similar to the needle-in-a-haystack test, when the context length to be tested is $128K$, it will be tested from $4K$ to $128K$ with $4K$ as the interval for a total of $N=32$ test data. For example, as shown in Figure~\ref{prompt}, when the context length is 96K, it will be tested from $3K$ to $96K$ with $3K$ as the interval for a total of $N=32$ test data.
		
		\begin{table*}[h!]
			\scriptsize
			\centering
			\renewcommand\tabcolsep{7pt}
			\renewcommand\arraystretch{1.5}
			
			\begin{tabular}{lcccccccc}
				\hline\hline
				\multirow{2}{*}{ \textbf{{Models}}} &  \multicolumn{2}{c}{\textsc{GPT-4 Turbo}} & \multirow{2}{*}{ {\textsc{Gemini 1.5 Pro}}}&  \multicolumn{3}{c}{\textsc{Claude3}} & \multirow{2}{*}{ {\textsc{GLM-4}}} & \multirow{2}{*}{ {\textsc{Moonshot-v1}}} \\ \cline{2-3} \cline{5-7}
				& \textit{1106} & \textit{0125} &  \textsc{} & \textsc{Opus} & \textsc{Sonnet} & \textsc{Haiku}  &  &  \\\hline
				\multicolumn{9}{l}{\textsc{P@32}} \\\hline
				Multi-evidence Searching (ZH) & \colorbox{myred}{0.697} & \colorbox{myred}{0.663} & \colorbox{myred}{0.775}& \colorbox{mygreen}{0.807}   & \colorbox{myred}{0.788} &  \colorbox{myred}{0.698} & \colorbox{myred}{0.682}& \colorbox{myred}{0.606} \\ 
				
				Multi-evidence Searching (EN) &  \colorbox{myred}{0.718} &  \colorbox{myred}{0.662} &  \colorbox{mygreen}{0.833}&  \colorbox{myred}{0.705}  &  -&- & \colorbox{myred}{0.389}&\colorbox{myred}{0.559}  \\ 
				\hline
				\multicolumn{9}{l}{\textsc{P@32$^\star$}} \\\hline
				Multi-evidence Reasoning (ZH) &  \colorbox{myred}{0.473} &  \colorbox{myred}{0.386} &  \colorbox{mygreen}{0.575}&  \colorbox{myred}{0.488} & -&- & \colorbox{myred}{0.475} & \colorbox{myred}{0.344}   \\ 
				Multi-evidence Reasoning (EN) & \colorbox{mygreen}{0.651} & \colorbox{myred}{0.610} & \colorbox{myred}{0.371}& \colorbox{myred}{0.374}  &-&- & \colorbox{myred}{0.179} & \colorbox{myred}{0.460}   \\ \hline

				Average Score & {0.635}$_{2}$ & 0.580$_{4}$ & 0.639$_{1}$ & 0.594$_{3}$  &  -&- & 0.431$_{6}$ & 0.492$_{5}$   \\ 
				
				\hline\hline
			\end{tabular}
			\caption{The overall performance on the Counting-Stars-(32)-(Multi-evidence Reasoning).}
			\label{overall}
		\end{table*}
		
		\begin{table*}[h!]
			\scriptsize
			\centering
			\renewcommand\tabcolsep{6pt}
			\renewcommand\arraystretch{1.45}
			\begin{tabular}{cccccccccccc}
				\hline\hline
				\multicolumn{1}{c}{\multirow{2}{*}{Models}}  & \multicolumn{11}{c}{Multi-evidence Searching (ZH)}                                                                  \\
				\multicolumn{1}{c}{}                 & 4K & 8K & 12K & 16K & 20K & 24K & 28K & 32K & 36K$\sim$64K & 68K$\sim$96K & 100K$\sim$128K\\\hline
				\textsc{GPT-4 Turbo} \textit{(1106)}             &  \colorbox{mygreen}{1.00}  &  0.97   &  0.92   &  0.76   &   0.84  &   0.75  &  0.75   &    0.78          &        0.76      &       0.61     &        0.57   \\
				\textsc{Claude3 Opus}                                &   \colorbox{mygreen}{1.00}  &   \colorbox{mygreen}{1.00}   &    0.89 &  0.83   &  0.86   &   0.89  & 0.85    &      0.78        &       0.78       &         \colorbox{mygreen}{0.76}         &    \colorbox{mygreen}{0.80}    \\
				\textsc{Gemini 1.5 Pro}                                &   \colorbox{mygreen}{1.00}  &   \colorbox{mygreen}{1.00}   &  0.97   &  \colorbox{mygreen}{0.91}   &   0.85  &  \colorbox{mygreen}{0.94}   &  0.61   &    0.68          &      0.80        &      0.74           &  0.67    \\
				
				\textsc{GLM-4}                                         &   \colorbox{mygreen}{1.00}  &  0.86   &   \colorbox{mygreen}{0.98}   &  0.90   &  \colorbox{mygreen}{0.96}   &  \colorbox{mygreen}{0.94}   &  \colorbox{mygreen}{0.88}   &      \colorbox{mygreen}{0.92}        &     \colorbox{mygreen}{0.84}          &      0.62      &    0.37    \\ 
				\textsc{Moonshot-v1}                               &  0.94  &  0.84   &   0.88  &  0.88   &  0.78   &   0.84  & 0.41    &      0.88        &       0.49       &     0.55          &  0.58       \\
				\hline\hline   
			\end{tabular}
			\caption{The \textsc{P@32} performance on the Chinese version of the Counting-Stars-(32)-(Multi-evidence Searching).}
			\label{mea-zh}
		\end{table*}

		\begin{table*}[h!]
			\scriptsize
			\centering
			\renewcommand\tabcolsep{2.5pt}
			\renewcommand\arraystretch{1.45}
			\begin{tabular}{ccccc|cccc|cccc}
				\hline\hline
				\multicolumn{1}{c}{\multirow{2}{*}{Models}} &  \multicolumn{4}{c|}{Multi-evidence Searching (EN)}       &  \multicolumn{4}{c|}{Multi-evidence Reasoning (ZH)}                            &  \multicolumn{4}{c}{Multi-evidence Reasoning (EN)}                                   \\
				\multicolumn{1}{c}{}                & 4K-32K & 36K-64K & 68K-96K & 100K-128K & 4K-32K & 36K-64K & 68K-96K & 100K-128K & 4K-32K & 36K-64K & 68K-96K & 100K-128K\\\hline
				\textsc{GPT-4 Turbo} \textit{(1106)}     &  {0.88}  &  0.62   &  0.64   &  0.74   &   0.80  &   0.51  &  0.28   &    0.30          &         \colorbox{mygreen}{0.86}      &        \colorbox{mygreen}{0.62}     &   \colorbox{mygreen}{0.60} & \colorbox{mygreen}{0.52}   \\
				\textsc{Claude3 Opus}     &  0.81  &   0.65   & 0.65 &  0.71   &   {0.82}   &  0.52  & 0.29    &     0.33     &       0.72       &     0.44  &  0.05 & 0.29    \\
				\textsc{Gemini 1.5 Pro}         &   \colorbox{mygreen}{0.90}  &   \colorbox{mygreen}{0.84}   &   \colorbox{mygreen}{0.80}   &  \colorbox{mygreen}{0.78}   &   0.75  &  0.55  &   \colorbox{mygreen}{0.50}   &     \colorbox{mygreen}{0.49}          &     0.66   &   0.24     & 0.30 & 0.28   \\
				
				\textsc{GLM-4}       &  0.57  &  0.36   &   0.34   & 0.29   & \colorbox{mygreen}{0.84} &  \colorbox{mygreen}{0.64} & 0.25 &   0.17  &     0.28     &     0.09      &   0.19  & 0.15  \\ 
				\textsc{Moonshot-v1}       &  0.86  &  0.43   &  0.45  &  0.50   &  0.65   &  0.24 & 0.19  &   0.29   &  0.52   &   0.53  &  0.44 &  0.35  \\
				\hline\hline   
			\end{tabular}
			\caption{The \textsc{P@32} performance on the English version of the Counting-Stars-(32)-(Multi-evidence Searching) as well as the Chinese and English versions of the Counting-Stars-(32)-(Multi-evidence Reasoning).}
			\label{left}
		\end{table*}

		\section{Experiments}
		
		\subsection{Baselines and Experimental Settings}
		In this study, we evaluate the Chinese and English versions of the Counting-Stars test on several famous long-context LLMs that may handle long contexts, including GPT-4 Turbo \cite{openai2024gpt4}, Gemini 1.5 Pro \cite{geminiteam2024gemini}, Claude 3 Opus\footnote{\url{https://www.anthropic.com/news/claude-3-family}}, GLM-4\footnote{\url{https://open.bigmodel.cn/}}, and Moonshot-v1\footnote{\url{https://kimi.moonshot.cn/}}.
		Table~\ref{ll} shows the context length limits (in tokens) of the LLMs GPT-4 Turbo, Gemini 1.5 Pro, Claude 3 Opus, GLM-4, and Moonshot-v1 used in the experiment.
		
		Specifically, in the experiments, we utilize the number of prompt tokens returned by the GPT-4 Turbo API to measure the context length. Therefore, it should also be noted that the position of inserting stars is somewhat biased. Firstly, it is due to the input context length being counted by the number of prompt tokens returned by GPT-4 Turbo. Secondly, it is precisely necessary to ensure some randomness. 
		
		Generally, evaluating text-based results is usually more complex, so the evidence to be collected in the Counting-Stars is all numerical, making it more straightforward to evaluate. For a piece of test data, the prediction results are evaluated starting from the first number of stars, that is, \textit{\{number1\}}, \textit{\{number2\}}, ..., \textit{\{numberM\}}. 
		In this paper, we adopt \textsc{P@N} as the evaluation metric, which includes two modes: one where \textsc{N=M}, representing the counting times (here, \textsc{M} denotes the counting times); the other, referencing prior studies \cite{keyphrase1, keyphrase2, keyphrase3, song2024countingstars}, adopts the \textsc{P@M} as the evaluation metric, where \textsc{M} denotes the total number of the retrieved results. 
		
		Specifically, for the Multi-evidence Searching task, if the results contain \textit{\{number1\}}, it gets a score of $1$; if it doesn't, it gets $0$. 
		Meanwhile, we construct a rule-based evaluation approach for the Multi-evidence Reasoning task, named \textsc{P@32$^\star$}. For \textsc{P@32$^\star$}, when the retrieved results contain only \textit{\{number1\}}, the score is $1$; if it also contains \textit{\{wrong number1\}}, the score is $0.5$; if the value only contains \textit{\{wrong number1\}}, the score is $0.25$, and if both If not found, the score is $0$. 
		
		\begin{figure*}[!h]
			\centering
			\includegraphics[scale=0.45]{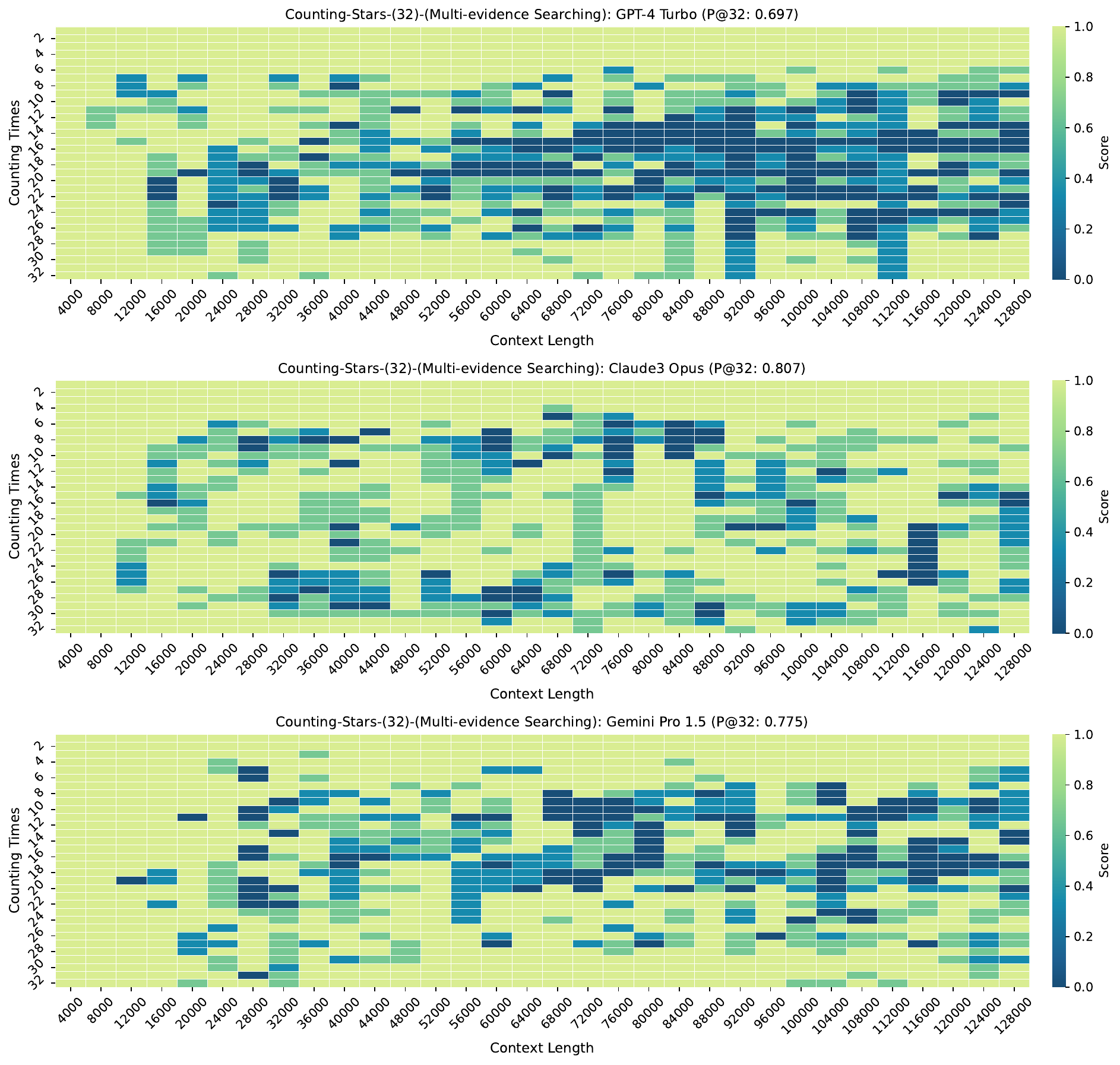}
			\caption{Visualization of the results on the Chinese version of the Counting-Stars-32-(Multi-evidence Searching).}
			\label{three_model}
		\end{figure*}
		\subsection{Overall Performance}
		Table~\ref{overall} present the performance of GPT-4 Turbo, Claude3 Opus, Gemini 1.5 Pro, GLM-4, Moonshot-v1 on the Chinese and English versions of the Counting-Stars-(32)-(Multi-evidence Searching) and Counting-Stars-(32)-(Multi-evidence Reasoning). Overall, Claude3 Opus achieves the best performance on the Chinese version of the Counting-Stars-(32)-(Multi-evidence Searching), Gemini 1.5 Pro obtains the best performance on the English version of the Counting-Stars-(32)-(Multi-evidence Searching) and the Chinese version of the Counting-Stars-(32)-(Multi-evidence Reasoning), and GPT-4 Turbo obtains the best performance on the English version of the Counting-Stars-(32)-(Multi-evidence Reasoning). Although these LLMs have achieved nearly perfect performance on the needle-in-a-haystack task, they still perform poorly on the Counting-Stars, which indicates that the needle-in-a-haystack is too simple to truly show the capabilities of LLMs in processing long texts.
		
\begin{figure*}[!h]
			\centering
			\includegraphics[scale=0.45]{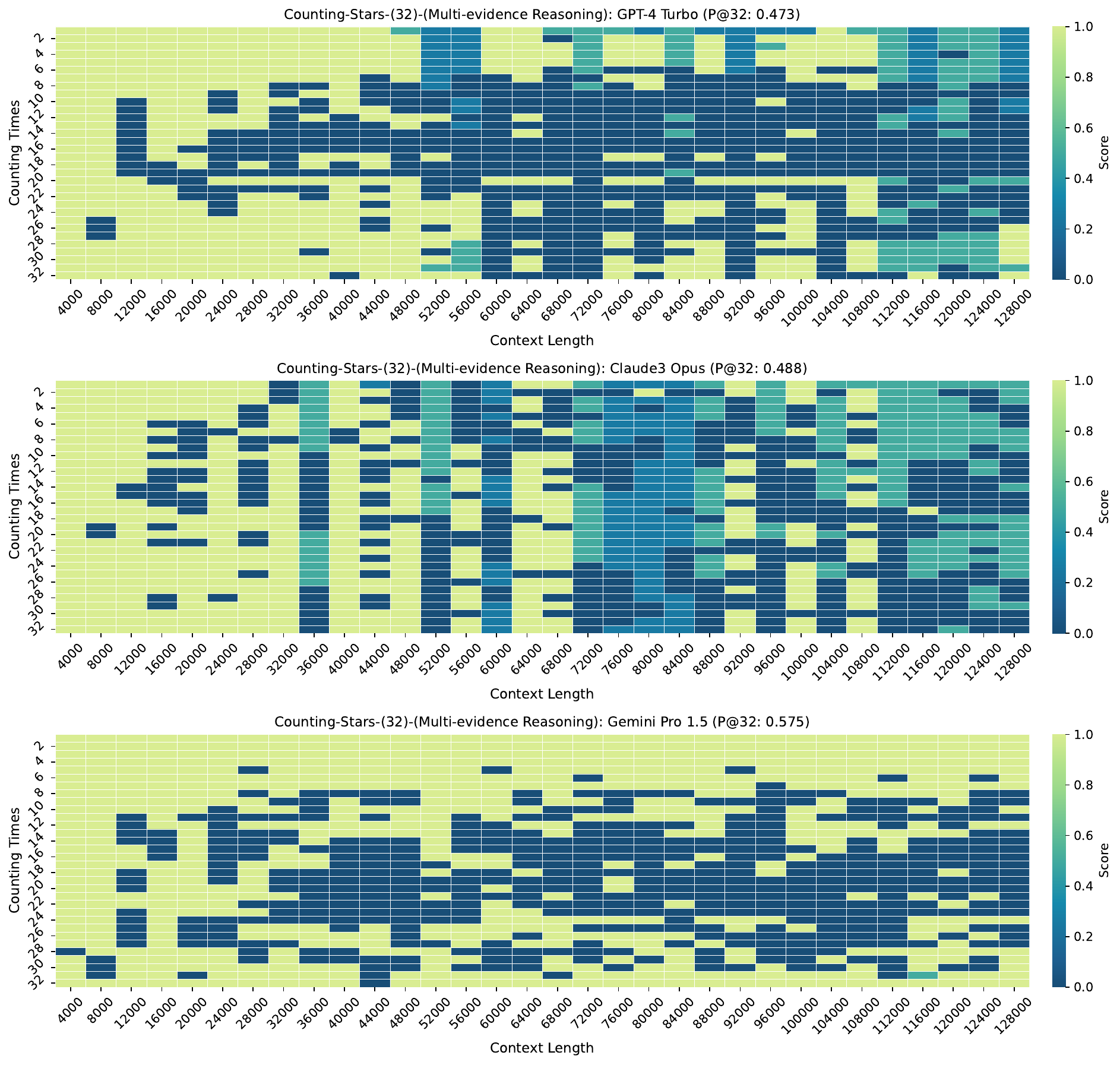}
			\caption{Visualization of the results on the Chinese version of the Counting-Stars-32-(Multi-evidence Reasoning).}
			\label{three_model_rea}
		\end{figure*}

		The multi-evidence reasoning task necessitates that LLMs engage in acquiring and reasoning multiple pieces of evidence simultaneously, which is more complex than the multi-evidence searching task. This task requires LLMs to sift through and exclude inaccurate evidence while gathering information from a long context to answer questions. As indicated by the data in Table~\ref{overall}, each LLM performs not well enough. Notably, in contrast to GPT-4 Turbo and Claude3 Opus, Gemini 1.5 Pro stands out for having gathered virtually no incorrect information, as shown in Figure~\ref{three_model_rea}.
		
		From Table~\ref{mea-zh} and Table~\ref{left}, it can be observed that all LLMs are capable of achieving higher scores in short-context scenarios, which confirms that the Counting-Stars is reasonable and can be accomplished by LLMs. However, as the context length increases, the performance of all models shows a downward trend. Among them, GPT-4 Turbo's performance is relatively stable.
		In addition, GLM-4 has obtained surprising results under the 32K context length of the Chinese version of the Counting-Stars-(32)-(Multi-evidence Searching). 
		
		By analyzing the experimental results of several long-context LLMs, we summarize three kinds of bad cases: (1) repeat a single number; (2) generate an increasing array; (3) fail to follow instruction, as shown in Table~\ref{bad_case}.

		\section{Discussion}
		We discuss the length-stability dilemma and the \textit{lost-in-the-middle} phenomenon in this section.
		
		\subsection{The Length-Stability Dilemma}
		One phenomenon that puzzles us the most among the test results of both needle-in-a-haystack and Counting-Stars is why the same task performs well when the input context length is long but badly at the shorter context (e.g., 112K and 108K in Figure~\ref{three_model}). It is important to note that this phenomenon becomes more pronounced as the length of the context increases. In other words, hiding the answer in different positions within different contexts results in LLMs failing to search it. Is this due to the different contexts surrounding the answer? Or is it because the distribution of the input context length of the training data is not uniform, leading to differences in the capabilities of LLMs across various context lengths? Therefore, could increase the robustness of LLMs help? 
		
		Based on the experiments in this paper, we are not yet able to determine the specific reasons behind this phenomenon; identifying these reasons is a goal that the next version of Counting-Stars aims to achieve. We consider the most intuitive explanation to be that the long-context capabilities of LLMs are still relatively weak, so when resources are limited, some stability must be sacrificed. Addressing this issue could help researchers better analyze and enhance the long-context modeling capabilities of LLMs, benefiting specific NLP tasks such as multi-document question answering. Moreover, stability refers to the understanding and reasoning abilities of LLMs when handling different long contexts, which is more crucial than merely the length of context processing.

		\begin{table*}[h!]
			\centering
			\scriptsize
				\resizebox{0.99\linewidth}{!}{
					\begin{tabular}[t]{@{}lp{0.8\linewidth}@{}}
						\toprule
						\textbf{Bad Case Description}  & \textbf{Example} \\
						\midrule
						\begin{tabular}[t]{@{}l@{}}repeat a single number \end{tabular} & 
						\begin{tabular}[t]{@{}p{\linewidth}@{}}
							[15, 15, 15, 15, 15, 15, 15, 15, 15, 15, 15, 15, 15, 15, 15, 15, 15, 15, 15, 15, 15, 15, 15, 15, 15, 15, 15, 15, 15, 15, 15, 15, 15, 15, 15, 15, 15, 15, 15, 15, 15, 15, 15, 15, 15, 15, 15, 15, 15, 15, 15, 15, 15, 15, 15, 15, 15, 15, 15, 15, 15, 15, 15, 15, 15, 15, 15, 15, 15, 15, 15, 15, 15, 15, 15, 15, 15, 15, 15, 15, 15, 15, 15, 15, 15, 15, 15, 15, 15, 15, 15, 15, 15, 15, 15, 15, 15, 15, 15, 15, 15, 15, 15, 15, 15, 15, 15, 15, 15, 15, 15, 15, 15, 15, 15, 15, 15, 15, 15, 15, 15, 15, 15, 15, 15, 15, 15, 15, 15, 15, 15, 15, 15, 15, ...]
							\\ 
						\end{tabular}
						\\
						\midrule
						\begin{tabular}[t]{@{}l@{}}generate increasing an array \end{tabular} & 
						\begin{tabular}[t]{@{}p{\linewidth}@{}}
							[5, 9, 15, 19, 29, 39, 45, 49, 53, 59, 63, 67, 71, 75, 79, 83, 87, 91, 95, 99, 103, 107, 111, 115, 119, 123, 127, 131, 135, 139, 143, 147, 151, 155, 159, 163, 167, 171, 175, 179, 183, 187, 191, 195, 199, 203, 207, 211, 215, 219, 223, 227, 231, 235, 239, 243, 247, 251, 255, 259, 263, 267, 271, 275, 279, 283, 287, 291, 295, 299, 303, 307, 311, 315, 319, 323, 327, 331, 335, 339, 343, 347, 351, ...]
							\\ 
						\end{tabular}
						\\
						\midrule
						\begin{tabular}[t]{@{}l@{}}fail to follow instruction \end{tabular} & 
						\begin{tabular}[t]{@{}p{\linewidth}@{}}
							"The little penguin counted 15 \Pisymbol{pzd}{72}", "The little penguin counted 117 \Pisymbol{pzd}{72}", "The little penguin counted 42 \Pisymbol{pzd}{72}", "The little penguin counted 69 \Pisymbol{pzd}{72}", "The little penguins counted 58 \Pisymbol{pzd}{72}", "The little penguin counted 107 \Pisymbol{pzd}{72}", "The little penguin counted 9 \Pisymbol{pzd}{72}", "The little penguin counted 49 \Pisymbol{pzd}{72}", "The little penguin counted 113 \Pisymbol{pzd}{72}",  …
							\\ 
						\end{tabular}
						\\
						\bottomrule
				\end{tabular}}
				\caption{Bad cases generated by LLMs.}
				\label{bad_case}
			\end{table*}
			
			\begin{table*}[h!]
				\scriptsize
				\centering
				\renewcommand\tabcolsep{16pt}
				\renewcommand\arraystretch{1.7}
				
				\begin{tabular}{lcccccc}
					\hline\hline
					\multirow{2}{*}{ \textbf{{Models}}} &  \multicolumn{2}{c}{\textsc{GPT-4 Turbo 1106}} & \multicolumn{2}{c}{\textsc{Gemini 1.5 Pro}} &  \multicolumn{2}{c}{\textsc{Claude3 Opus}}  \\ \cline{2-7}
					& \textit{F1@32} & \textit{F1@M} & \textit{F1@32} & \textit{F1@M} & \textit{F1@32} & \textit{F1@M}  \\\hline
					Multi-evidence Searching (ZH) & 0.808 & 0.808 & 0.866 &  0.866 & 0.887 & 0.889 \\ 
					
					Multi-evidence Searching (EN) & 0.789 &  0.790 & 0.904 & 0.905  &  0.784 & 0.784  \\ 
					\hline
					Multi-evidence Reasoning (ZH) & 0.597 & 0.601 & 0.719 & 0.719 & 0.606 & 0.634  \\ 
					Multi-evidence Reasoning (EN) & 0.757 & 0.769 & 0.373 & 0.404 & 0.457 & 0.468  \\ 
					\hline\hline
				\end{tabular}
				\caption{The \textsc{F1@32} and \textsc{F1@M} performance on the Counting-Stars.}
				\label{overall}
			\end{table*}
			
			\subsection{Lost in the Middle}
			Prior research indicates a performance decline in some LLMs when answers are positioned around the middle of the long context \cite{lost}. Similar to \cite{inftybench}, however, our findings can not strongly corroborate the \textit{lost-in-the-middle} phenomenon. One possible reason why we obtain different observations from \cite{lost} is that they find the phenomenon via the test at most 16K length contexts, which is not long enough. In our experiments based on the Counting-Stars, we discover that the bad cases may not mainly appear in the middle of the long context, especially for the results of Claude3 Opus, as shown in Figure~\ref{three_model}. Hence, we hypothesize that the \textit{lost-in-the-middle} phenomenon only occurs in specific tasks, length contexts, or models. 
			
			By observing the results of multiple experiments, we guess that the \textit{lost-in-the-middle} phenomenon of LLMs is determined by their implicit reasoning or thinking patterns when dealing with specific tasks or length contexts. Interestingly, as illustrated in Figure~\ref{bad_case} ("fail to follow instruction"), when collecting the numbers of stars, LLMs first attempt to memorize and recite relevant sentences and then further summarize them into the final result. According to the above findings, we guess this kind of implicit reasoning or thinking pattern may alleviate the \textit{lost-in-the-middle} phenomenon.
			
			\section{Related Work}
			Prior research on long-context modeling has traditionally adopted perplexity as the primary evaluation metric \cite{peng, LongLoRA}. Meanwhile, synthetic tasks (e.g., retrieval tasks) have been employed to gauge the capacity of LLMs to handle extremely long inputs \cite{Li}. However, as highlighted in \citet{Xiong}, neither perplexity scores nor performance on synthetic tasks may fully capture the effectiveness of LLMs in real-world applications. Several benchmarks proposed by \citet{longbench, L-Eval, LV-Eval, clongeval, inftybench} recently aim to evaluate long-context LLMs. 
			
			A recent benchmark for testing the long-context LLMs is needle-in-a-haystack, which asks LLMs to recite the information in a “needle” sentence (“The best thing to do in San Francisco is eat a sandwich and sit in Dolores Park on a sunny day”) that is inserted at a designed location in a long text. The difference between the needle-in-a-haystack and existing benchmarks is that it does not rely on specific data, especially those that may be utilized to train LLMs. In addition, the needle-in-a-haystack can be treated as a benchmark where the test data can be easily replaced to mitigate the issue of data leakage, which generally occurs in existing long-context benchmarks. As mentioned before, however, many recently released LLMs evaluate the capability of long-context handling by testing the needle-in-a-haystack, all achieving nearly perfect performance, making it impossible to distinguish the gaps between different long-context LLMs.
			
			The Counting-Stars evaluates the capabilities of multi-evidence searching and reasoning of LLMs, which should be more noteworthy in the long context modeling of LLMs, as reflected in tasks such as multi-document question answering and summarization. 
			Concretely, the former primarily evaluates the capability of LLMs to collect multiple pieces of evidence simultaneously (distinguishing between long context and inserted sentences), while the latter tests the ability of LLMs to gather and reason various pieces of evidence at the same time correctly, that is, reasoning is required when collecting information (distinguishing between correct and incorrect evidence in inserted sentences). \textit{To the best of our knowledge, the Counting-Stars is the first scalable long-context benchmark to ask LLMs to simultaneously differentiate between correct and incorrect evidence in each inserted sentence}.
			
			Furthermore, similar to the recent benchmark \cite{LooGLE}, we refer to the long-dependency tasks as those that require capturing and understanding inter-dependency across multiple pieces of evidence spanning the entire long context. Hence, the Counting-Stars can also be considered a long-dependency task when calculating scores from the sample level, i.e., one testing sample only computes one score. In addition, since sentences are pieces of evidence and distributed throughout the entire long context, it is expected that other abilities behind long-context LLMs could be analyzed, including the long-context processing strategies and attention mechanisms, which is meaningful for studying the capability of long-context LLMs.
			It is worth mentioning that the cost of the Counting-Stars is lower than that of the needle-in-a-haystack, which is beneficial for reducing carbon emissions.
			
			\section{Conclusion}
			In this paper, we propose Counting-Stars, a multi-evidence, position-aware, and scalable benchmark for evaluating the multiple pieces of evidence retrieval capabilities of LLMs via two counting-based tasks. Utilizing the Counting-Stars, we conduct intriguing analyses on the behavior of LLMs, including the length-stability dilemma and the absence of the "lost-in-the-middle" phenomenon. Our analysis provides valuable insights into how LLMs handle long contexts, which can inform and guide future research endeavors.

			\section{Limitations}
			While this paper offers some insights into the performance of long-context LLMs, it may not be sufficiently diverse or extensive to provide a comprehensive evaluation of the long-context capabilities of LLMs, a constraint common to most analyses and benchmarks. However, we consider that for long-context scenarios, the capability to acquire multi-evidence is the most critical capability, which is also the central aspect tested by the Counting-Stars. Finally, we analyze and summarize potential uncertainty in our experiments, experiments on more famous LLMs.
			
			\noindent (1) \textbf{Potential uncertainty in our experiments.} 
			
			\noindent \textit{(a) Context Used}. 
			
			Through our experiments, it has been discovered that for tests like needle-in-a-haystack and Counting-Stars, different contexts may cause variations in the results. However, all experiments use the same context information in this paper. It must be noted, though, that different LLMs show variations in performance across different contexts.

			\noindent \textit{(b) Prompt Used}. 
			
			It is well known that LLMs are very sensitive to the design of prompts, and the results of different prompts can vary greatly. However, our experiments only constructed reasonable prompts that clearly express the task requirements without deliberately optimizing the prompts. From the experimental results, each model understood the prompts correctly without ambiguity.

			\noindent \textit{(c) Service Used}. 
			
			Due to regional access restrictions on the tested LLMs in this paper, two different services are used to test the five LLMs discussed in this paper: API and poe.com. Concretely, the latter approach does not allow adjusting model parameters, such as temperature. Therefore, when using the former, we set the temperature to 0 to ensure, as much as possible, the fairness of evaluation settings. However, using different access approaches may introduce some hard-to-find issues, potentially leading to biases in the testing results.
			
			\noindent \textit{(d) Evaluation Used}.
			
			Actually, the adopted evaluation metric in this paper seems too simple, particularly in the multi-evidence reasoning task. A more comprehensive and reasonable evaluation metric may better reflect LLMs' long-context capability, such as using different context lengths as weights.
			
			\noindent (2) \textbf{Experiments on more LLMs.} 
			
			\noindent In the future, we will evaluate more famous LLMs on the Counting-Stars, such as Llama3\footnote{\url{https://github.com/meta-llama/llama3}}, Mistral\footnote{\url{https://mistral.ai/news/la-plateforme/}}, Llama2 \cite{touvron2023llama}, Mixtral \cite{jiang2024mixtral}, Command-R\footnote{\url{https://docs.cohere.com/docs/command-r\#model-details}}, LongLoRA \cite{chen2024longlora}, LongAlpaca\cite{chen2024longlora}, LWM \cite{liu2024world}, Qwen \cite{bai2023qwen}, DeepSeek-V2 \cite{deepseekai2024deepseekv2}, etc.
			
			\noindent (3) \textbf{Future expansion of the Counting-Stars.} 
			
			\noindent Initially, the Counting-Stars is designed to require LLMs to count the total number of stars in all sentences inserted in the long context, which aims to test the multi-evidence searching of LLMs from a long dependency perspective. However, we find that if LLMs are required to calculate the total number of stars, they usually perform poorly. Specifically, we analyze the reasons for the bad performance, which mainly include three points: 
			\begin{itemize}
				\item LLMs are unable to discover the sentences describing the little penguin counted stars.
				\item LLMs are able to discover all sentences but cannot remember them all.
				\item LLMs are able to remember all sentences but need better mathematical ability to calculate the total numbers correctly.
			\end{itemize}
			Still, we find that even if it is a simple mathematical problem of calculating “1 + 1 + 1 + 1 + 1 + 1 + 1 + 1 + 1 + 1 + 1 + 1 + 1 + 1 + 1 + 1 + 1  + 1 + 1 + 1 + 1 + 1 + 1 + 1 + 1 + 1 + 1 + 1 + 1 + 1 + 1 + 1 + 1 + 1 + 1”, the probability of LLMs calculating correctly is lower. So, introducing the summation operation may be a simple and direct extension of the Counting-Stars. However, this paper currently focuses on the multi-evidence retrieval abilities of LLMs in long contexts. Therefore, we have chosen to require LLMs to list all the numbers of stars.
			
			To further optimize and enhance the Counting-Stars, we imagine other evaluation strategies similar to the Counting-Stars, such as scattering stars by different players and specifying LLMs to search for the stars counted by one of them. Based on the above idea, adding more complex interactions between players may construct a more difficult question for evaluating LLMs.
			
			\section*{Acknowledgments}
			We thank the three anonymous reviewers for carefully reading our paper and their insightful comments and suggestions.
			\bibliography{custom}
		\end{document}